\documentclass{article} 
\usepackage[final]{colm2026_conference_arxiv}

\usepackage[utf8]{inputenc} 
\usepackage[T1]{fontenc}    
\usepackage{hyperref}       
\usepackage{url}            
\usepackage{booktabs}       
\usepackage{lineno}
\usepackage{nicefrac}       
\usepackage{microtype}      
\usepackage{graphicx}       
\usepackage{multirow}       
\usepackage{colortbl}       
\usepackage{xcolor}         
\usepackage{subcaption}     
\usepackage[inline]{enumitem}
\usepackage{enumitem}
\usepackage{boldline}
\usepackage{hhline}
\usepackage{wrapfig}
\usepackage{float}

\usepackage{amsmath,amsfonts,bm}

\def\eqref#1{equation~\ref{#1}}

\def\1{\bm{1}}

\DeclareMathAlphabet{\mathsfit}{\encodingdefault}{\sfdefault}{m}{sl}
\SetMathAlphabet{\mathsfit}{bold}{\encodingdefault}{\sfdefault}{bx}{n}

\definecolor{oursrow}{RGB}{232,243,255}

\title{Learning While Acting: A Skill-Enhanced Test-Time Co-Evolution Framework for Online Lifelong Learning Agents}

\author{
  \textbf{Bo Mao}$^{1}$, Jie Zhou$^{1,2}$\footnotemark[1], \textbf{Yutao Yang}$^1$, \textbf{Xin Li}$^2$, \textbf{Xian Wei}$^3$, \textbf{Qin Chen}$^1$, \textbf{Xingjiao Wu}$^1$, \textbf{Liang He}$^1$ \\
  $^1$ School of Computer Science and Technology, East China Normal University, \\
   $^2$ Shanghai AI Laboratory, \\ 
   $^3$ Software Engineering Institute, East China Normal University \\ 
  \texttt{\{jzhou, qchen, lhe\}@cs.ecnu.edu.cn}
}

\begin{document}

\ifcolmsubmission
\linenumbers
\fi

\maketitle

\begin{abstract}
Lifelong learning is essential for Large Language Model (LLM) agents operating in dynamic, interactive environments. However, existing lifelong learning agents for long-horizon tasks typically depend on discrete skill or past experiences retrieval with static parameters during inference, which prevents them from continuously internalizing test-time feedback like human learners. To bridge this gap, we propose Skill-enhanced Test-Time Co-Evolution (\texttt{LifeSkill}), a two-stage reinforcement learning framework for Online Lifelong Learning Agents. 
Specifically, we design Verifier-Guided Skill Learning that addresses the lack of direct supervision for skill extraction by rewarding candidate skills according to the average verifier success of multiple skill-conditioned policy rollouts, encouraging the model to generate skills that are useful for solving tasks rather than merely plausible in text. 
Furthermore, we introduce Online Skill Internalization, which continuously improves the policy model during test-time interaction by transforming skill-conditioned trajectories into reward signals. 
This enables the agent to directly internalize reasoning capabilities into its parameters, avoiding the context bloat of experience retrieval. 
Experiments on LifelongAgentBench show that LifeSkill improves average performance by 7 absolute points by comparing with existing lifelong agent baselines.
\end{abstract}

\section{Introduction}
\label{sec:intro}

Large language model (LLM) agents have shown promising performance in long-horizon decision making, interactive planning, and sequential problem solving. However, real-world environments are dynamic: task distributions evolve over time, useful strategies emerge through repeated interaction, and agents must continually accumulate knowledge from experience. This makes lifelong learning or continual learning a core requirement for LLM agents \citep{gao2025survey,yang2025recent}. Rather than treating each episode as an isolated inference problem, a lifelong agent should continuously improve while interacting with the environment, converting test-time feedback into persistent capability growth \citep{fang2025comprehensive,cai2025building}.

Recent studies have taken initial steps toward this goal, mainly through memory-based experience reuse and test-time self-improvement. Memory-driven agents such as Synapse \citep{DBLP:conf/iclr/ZhengWW024}, AWM \citep{DBLP:conf/icml/WangMFN25}, AgentKB \citep{DBLP:journals/corr/abs-2507-06229}, and CER \citep{DBLP:conf/acl/LiuSNY25} store and retrieve past trajectories or summarized workflows to guide future decisions. In parallel, test-time scaling and self-reflection methods improve performance by exploring multiple reasoning paths and revising failed attempts during inference \citep{DBLP:conf/nips/ShinnCGNY23,DBLP:conf/nips/MadaanTGHGW0DPY23}. Reinforcement learning with verifiable rewards has further shown strong potential for improving reasoning in domains with deterministic outcomes \citep{DBLP:journals/corr/abs-2501-12948}. Despite these advances, existing lifelong learning agents still fall short of genuine continual adaptation.

In particular, current approaches face three key limitations. \textbf{First}, most agents keep model parameters fixed at test time and rely on external memories to improve performance, which prevents them from truly learning during deployment. \textbf{Second}, although prior work can summarize trajectories or produce reflections, how to extract \emph{reusable skills} from interaction feedback remains insufficiently explored, especially under reliable and execution-grounded supervision. \textbf{Third}, acquired knowledge is usually represented as discrete external text, making it difficult for the model to internalize, compose, and generalize these skills as part of its parametric knowledge. As a result, existing agents can retrieve experience, but cannot effectively transform it into lasting capability.

\begin{figure}[t]
\vspace{-4mm}
    \centering
    \includegraphics[width=0.85\textwidth]{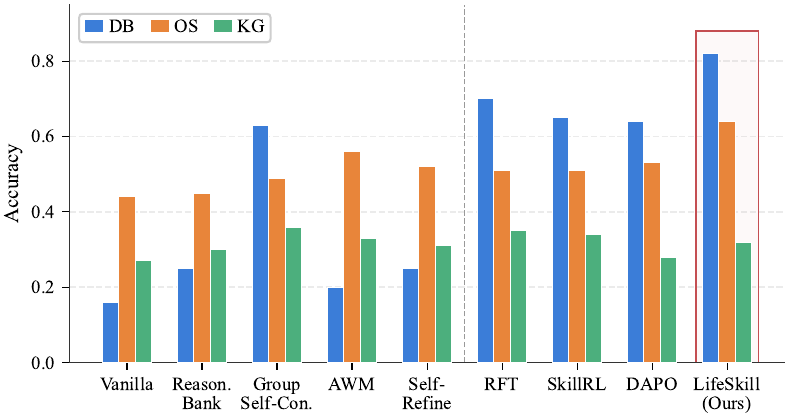}
    \vspace{-2mm}
    \caption{Comparison of all methods on LifelongAgentBench.  Left of the dashed line: training-free (memory retrieval) baselines; right: training-based methods.  \texttt{LifeSkill} (highlighted) achieves the highest accuracy on DB and OS and the best overall average.}
    \label{fig:main_results}
    \vspace{-5mm}
\end{figure}

To address these challenges, we propose \texttt{LifeSkill}, a Skill-Enhanced Test-Time Co-Evolution Framework for online lifelong LLM agents. The key idea is to use reinforcement learning to improve both \emph{what skills the agent extracts from failure} and \emph{how those skills are absorbed into the policy}. In the first stage, Verifier-Guided Skill Learning addresses the lack of direct supervision for skill extraction. After a failed attempt, the skill extractor proposes candidate skills, and each skill is evaluated by the average verifier reward of multiple skill-conditioned policy rollouts. This trains the extractor to generate skills that are useful for solving tasks, rather than merely plausible in text. In the second stage, Online Skill Internalization addresses the mismatch between skill-guided exploration and deployment. We remove explicit skill text from successful skill-conditioned trajectories and update the policy under the original task input alone, so that behaviors discovered during interaction are internalized into model parameters instead of remaining as external memory. Together, these two stages form a test-time co-evolution loop: better skill extraction leads to more effective exploration, while improved policy parameters in turn yield stronger trajectories for subsequent skill learning.

We evaluate \texttt{LifeSkill} on LifelongAgentBench, a benchmark for continual adaptation in long-horizon interactive tasks. Experimental results (Figure~\ref{fig:main_results}) show that \texttt{LifeSkill} consistently outperforms strong baselines and improves average performance by 7 absolute points. Further analysis confirms that both verifier-guided skill learning and online skill internalization are critical for sustained lifelong adaptation.

Our main contributions are as follows:
\begin{itemize}[leftmargin=*, align=left]
    \item We propose \texttt{LifeSkill}, a two-stage reinforcement learning framework that enables online lifelong LLM agents to improve during deployment, moving beyond static-parameter inference with external memory retrieval.
    \item We introduce Verifier-Guided Skill Learning to train a skill extractor from execution-grounded feedback, and Online Skill Internalization to convert successful skill-guided trajectories into unscaffolded policy improvements.
    \item Extensive experiments on LifelongAgentBench show that \texttt{LifeSkill} consistently outperforms strong baselines, improving average performance by 7 absolute points.
\end{itemize}

\section{Related Work}
\label{sec:related_work}

\textbf{Self-evolving LLM Agents.}
Recent surveys and benchmarks argue that agent capability should be studied as \emph{continual adaptation} rather than single-episode inference \citep{gao2025survey,fang2025comprehensive,yang2025recent,cai2025building,wang2024survey_agents,liu2023agentbench,zhou2023webarena,koh2024visualwebarena,zheng2025lifelongagentbenchevaluatingllmagents}. A dominant line augments agents with external memory or experience banks, including trajectory exemplars, workflow memories, contextual replay buffers, cross-domain knowledge bases, and persistent memory systems \citep{DBLP:conf/iclr/ZhengWW024,DBLP:conf/icml/WangMFN25,DBLP:conf/acl/LiuSNY25,DBLP:journals/corr/abs-2507-06229,DBLP:journals/corr/abs-2509-25140,packer2024memgpt,li2025memos,suzgun2025dynamic}. Related agents also accumulate reusable reflections or executable routines in libraries \citep{wang2023voyager,zhao2024expel,park2023generative}. These approaches show that experience reuse is valuable, but most gains remain \emph{outside} model parameters and therefore incur retrieval overhead, context growth, or memory interference at deployment time.

\textbf{Test-time Learning.}
A complementary direction improves performance by allocating more inference-time computation to decomposition, revision, and search \citep{wei2022cot,wang2023selfconsistency,zhou2023leasttomost,yao2023react,DBLP:conf/nips/ShinnCGNY23,DBLP:conf/nips/MadaanTGHGW0DPY23,yao2023tree,besta2024graph,zhou2023lats}. Such methods can recover from local mistakes and better handle long-horizon tasks by exploring multiple candidate reasoning paths or action plans. However, their gains are usually transient: the agent often repeats similar search or self-correction on related future tasks because the discovered strategies are not persistently absorbed into the policy. Our work keeps the benefits of test-time exploration, but treats it as supervision for future capability rather than only extra computation for the current episode.

\textbf{Skill Abstraction.}
A third line of work represents reusable behavior as skills, tools, programs, or temporally extended options. Temporal abstraction in reinforcement learning formalizes reusable high-level behaviors \citep{sutton1999options}, while LLM agents increasingly externalize procedures as code modules, tool calls, or workflow descriptions \citep{wang2023voyager,zhao2024expel,schick2023toolformer,gao2023pal,chen2023pot}. These formulations improve compositionality and interpretability, but most prior work either assumes manually designed tool interfaces, relies on offline accumulation of successful routines, or keeps the discovered skills as explicit scaffolds during future inference. In contrast, we treat skills as an intermediate learning interface: they guide exploration after failure and then internalize the resulting behavior back into the policy.

\textbf{Online Continual Learning.}
At the parameter level, continual learning studies how to retain prior knowledge while acquiring new behavior, using mechanisms such as regularization, replay, and online sample selection \citep{kirkpatrick2017ewc,schwarz2018progresscompress,rolnick2019experience,aljundi2019mir}. Test-time adaptation further updates models during deployment under distribution shift \citep{sun2020ttt,wang2021tent}. For LLM post-training, instruction alignment and reasoning self-improvement have been advanced by policy optimization, self-training, and verifier-based learning \citep{ouyang2022instructgpt,schulman2017ppo,zelikman2022star,DBLP:journals/corr/abs-2308-01825,DBLP:journals/corr/abs-2501-12948,DBLP:journals/corr/abs-2503-14476}. Meanwhile, judge-based reflection pipelines often rely on subjective LLM evaluators \citep{DBLP:conf/nips/ZhengC00WZL0LXZ23}. Our setting differs in two aspects: adaptation must happen online inside interactive environments, and supervision must come from task verifiers rather than preference labels or free-form judges.

Overall, \texttt{LifeSkill} differs from prior lifelong agent methods in three key ways. First, it learns which skills to extract from failures using execution-grounded verifier feedback, rather than heuristic summaries or judge-scored reflections. Second, it performs online parameter updates during deployment instead of relying solely on external memory retrieval. Third, it internalizes successful skill-guided behaviors under the original task input, reducing long-term dependence on retrieved text.

\begin{figure}[t]
    \centering
    \includegraphics[width=\textwidth]{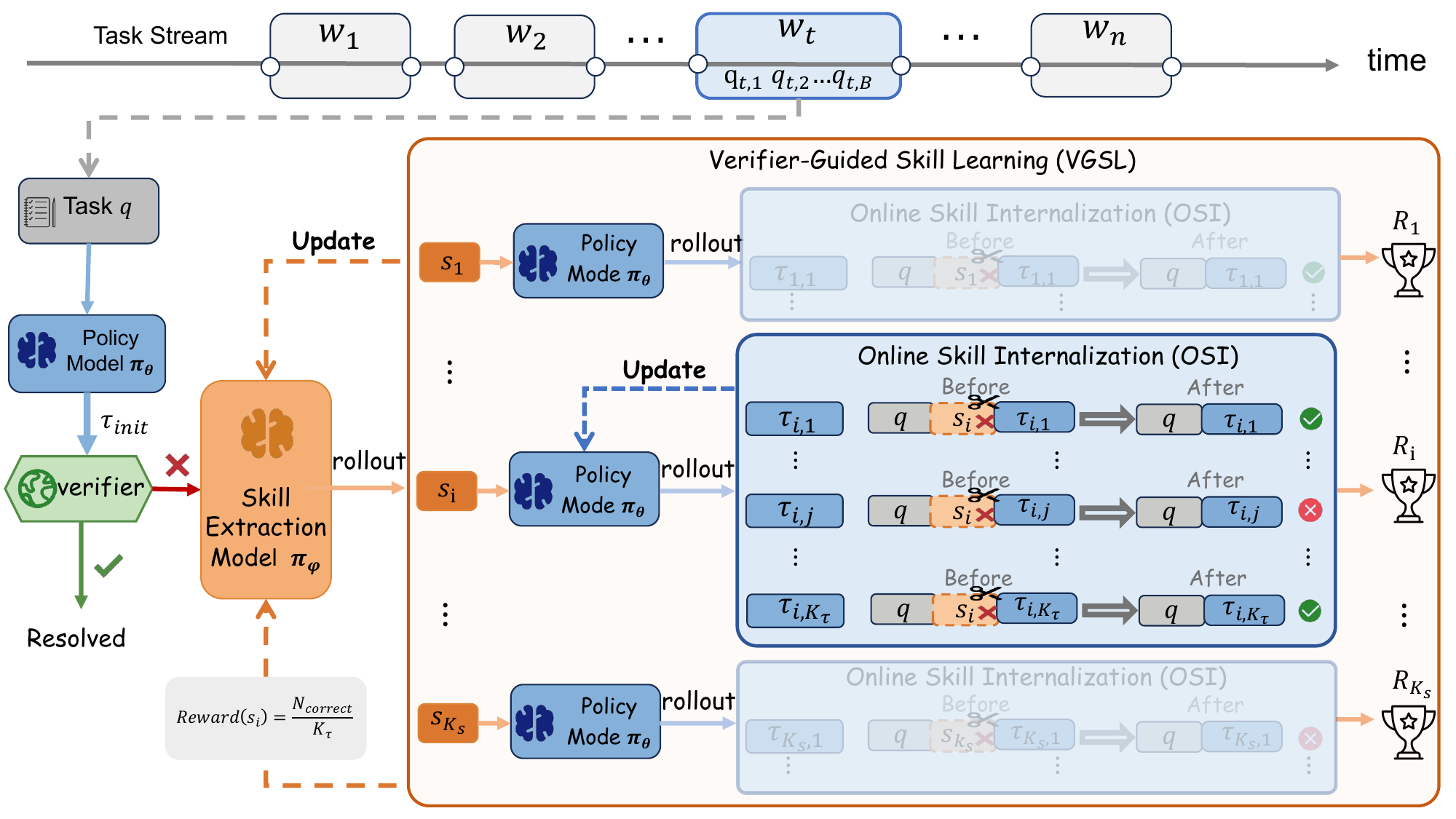}
    \vspace{-3mm}
    \caption{Overview of \texttt{LifeSkill}. 
    Verifier-Guided Skill Learning optimizes the extractor using the average verifier success of these skill-conditioned executions, while Online Skill Internalization removes explicit skill text from verified successful trajectories and updates the Policy Model under the original task input alone. 
    }
    \label{fig:method}
    \vspace{-4mm}
\end{figure}

\vspace{-1mm}
\section{Our \texttt{LifeSkill} Method}
\label{sec:methodology}
\vspace{-2mm}
We study online lifelong learning over a stream of tasks and instantiate \texttt{LifeSkill} as a skill-enhanced test-time co-evolution framework. Unlike approaches that keep model parameters fixed during deployment or treat adaptation as a separate post-hoc step, \texttt{LifeSkill} learns while interacting with the current task stream. It maintains two models with the same backbone architecture but different parameters: a \emph{Policy Model} $\pi_\theta$ for task execution and a \emph{Skill Extraction Model} $\pi_\phi$ for distilling reusable skills from failed interactions. The two models are coupled through two reinforcement learning mechanisms. Verifier-Guided Skill Learning (VGSL; \S\ref{subsec:vgsl}) trains $\pi_\phi$ using downstream verifier outcomes, encouraging it to extract skills that improve execution rather than merely sounding plausible. Online Skill Internalization (OSI; \S\ref{subsec:osi}) updates $\pi_\theta$ online using successful skill-conditioned trajectories after removing the explicit skill scaffold from the input, so that useful behaviors are absorbed into the model parameters. Figure~\ref{fig:method} illustrates the overall online co-evolution loop.

\subsection{Problem Formulation: Online Lifelong Learning}
\label{subsec:problem}

We consider an online stream of tasks
\begin{equation}
    \mathcal{D}_{\text{stream}} = (q_1, q_2, \ldots).
\end{equation}
Following prior lifelong evaluation protocols, we process the stream using non-overlapping windows. At step $t$, the agent receives a batch
\begin{equation}
    \mathcal{W}_t = \{q_{t,1}, \ldots, q_{t,B}\},
\end{equation}
where $B$ is the window size. For a task $q$ and an executable trajectory $\tau$, the environment returns a deterministic binary verdict
\begin{equation}
    r = \mathcal{V}(q, \tau) \in \{0,1\},
\end{equation}
where $\mathcal{V}$ is the task-specific verifier and $r=1$ indicates success.

The agent must solve the tasks in $\mathcal{W}_t$ and update its models before moving to $\mathcal{W}_{t+1}$. During this process, it can only access the current and past windows, without future tasks, human annotations, or judge-model supervision. The goal is to maximize cumulative verified success over the task stream:
\begin{equation}
    \max \sum_t \sum_{q \in \mathcal{W}_t} \mathbb{E}\left[\mathcal{V}(q, \tau_q)\right].
\end{equation}

\subsection{Overview of LifeSkill}
\label{subsec:online_loop}

\texttt{LifeSkill} operates in a continual online loop over the task stream. For each task $q \in \mathcal{W}_t$, the Policy Model first attempts the task under the original input:
\begin{equation}
    \tau^{(0)} \sim \pi_\theta(\cdot \mid q),
\end{equation}
with verifier reward
\begin{equation}
    r^{(0)} = \mathcal{V}(q, \tau^{(0)}).
\end{equation}
If $r^{(0)} = 1$, the task is solved. Otherwise, the Skill Extraction Model summarizes reusable guidance from the failed attempt:
\begin{equation}
    s \sim \pi_\phi(\cdot \mid q, \tau^{(0)}),
\end{equation}
and the Policy Model performs an additional skill-conditioned rollout:
\begin{equation}
    \tau^{(1)} \sim \pi_\theta(\cdot \mid q, s), \qquad r^{(1)} = \mathcal{V}(q, \tau^{(1)}).
\end{equation}

These interactions serve two roles simultaneously. On the one hand, skill-conditioned retries improve current-window performance by enabling targeted exploration after failure. On the other hand, the collected verifier signals provide online supervision for updating both models before the agent moves to $\mathcal{W}_{t+1}$. Specifically, VGSL improves the quality of extracted skills, while OSI turns successful skill-guided behavior into unscaffolded parametric improvements. In this way, acting, skill discovery, and parameter adaptation are integrated into one continuous online loop.

\subsection{Verifier-Guided Skill Learning}
\label{subsec:vgsl}

A key difficulty in training the Skill Extraction Model is that skill descriptions are free-form text and therefore lack explicit ground-truth labels. A common workaround is to score them with an auxiliary LLM judge, but such supervision is subjective and only indirectly related to task success. Instead, we supervise $\pi_\phi$ using \emph{verifiable} downstream outcomes from online execution. Since the extractor outputs a variable-length skill sequence rather than a discrete label, we optimize it with the standard DAPO objective \citep{DBLP:journals/corr/abs-2503-14476}.

Consider a failed initial attempt $(q, \tau^{(0)})$ with $\mathcal{V}(q, \tau^{(0)}) = 0$. VGSL evaluates candidate skills by their effect on downstream success. First, the Skill Extraction Model samples $K_s$ candidate skills:
\begin{equation}
    s_i \sim \pi_\phi(\cdot \mid q, \tau^{(0)}), \qquad i = 1, \ldots, K_s.
\end{equation}
For each candidate skill $s_i$, the Policy Model then samples $K_\tau$ skill-conditioned trajectories:
\begin{equation}
    \tau_{i,j} \sim \pi_\theta(\cdot \mid q, s_i), \qquad j = 1, \ldots, K_\tau,
\end{equation}
and obtains verifier rewards:
\begin{equation}
    r_{i,j} = \mathcal{V}(q, \tau_{i,j}).
\end{equation}
We define the utility of skill $s_i$ by its empirical success rate:
\begin{equation}
    R_i \;=\; \frac{1}{K_\tau} \sum_{j=1}^{K_\tau} r_{i,j}.
\end{equation}
Intuitively, $R_i$ measures whether the skill improves executable behavior, rather than whether it merely appears plausible in text. We then use these verifier-based skill utilities to optimize the Skill Extraction Model with DAPO. Concretely, for each failed attempt, the sampled candidate skills form a comparison group, and we convert their utilities into normalized relative advantages:
\begin{equation}
    \hat{A}_i \;=\; \frac{R_i - \mu_R}{\sigma_R + \epsilon},
    \qquad
    \mu_R \;=\; \frac{1}{K_s}\sum_{i=1}^{K_s} R_i,
    \qquad
    \sigma_R \;=\; \sqrt{\frac{1}{K_s}\sum_{i=1}^{K_s}(R_i-\mu_R)^2}.
\end{equation}
For a tokenized skill $s_i = (s_{i,1}, \ldots, s_{i,|s_i|})$, we define the token-level importance ratio as
\begin{equation}
    \rho_{i,\ell}(\phi)
    \;=\;
    \frac{\pi_\phi(s_{i,\ell} \mid q, \tau^{(0)}, s_{i,<\ell})}
    {\pi_{\phi_{\mathrm{old}}}(s_{i,\ell} \mid q, \tau^{(0)}, s_{i,<\ell})}.
\end{equation}
The Skill Extraction Model is then updated with the DAPO token-level clipped objective:
\begin{equation}
    \mathcal{L}_{\text{VGSL}}(\phi)
    \;=\;
    -\mathbb{E}_{i,\ell}
    \left[
    \min\left(
    \rho_{i,\ell}(\phi)\hat{A}_i,\,
    \operatorname{clip}\!\big(
    \rho_{i,\ell}(\phi),\, 1-\epsilon_{\text{low}},\, 1+\epsilon_{\text{high}}
    \big)\hat{A}_i
    \right)
    \right].
\end{equation}
Here, the expectation is taken over all generated skill tokens for failed tasks in the current window. In this way, VGSL preserves the optimization stability of DAPO while grounding the learning signal in downstream verified execution outcomes, so the extractor is rewarded for proposing skills that actually improve policy rollouts.


\subsection{Online Skill Internalization}
\label{subsec:osi}

VGSL improves online exploration by discovering useful skill text, but an online lifelong agent must also convert these discoveries into persistent capability improvements during deployment. Otherwise, useful skills would remain external guidance that needs to be regenerated or reintroduced repeatedly, rather than becoming part of the policy itself.

Let $\mathcal{B}_t = \{(q, s, \tau, r)\}$
denote the collection of all skill-conditioned trajectories gathered in window $t$, including both the online retries and the rollouts used by VGSL.

For each sample $(q, s, \tau, r) \in \mathcal{B}_t$, we remove the explicit skill from the conditioning context and retain only the original task-query/output pair:
\begin{equation}
    (q, s, \tau) \;\Rightarrow\; (q, \tau).
\end{equation}
Importantly, we do not modify the output trajectory itself; we only remove the external scaffold from the input.

Because the conditioning context changes from $(q, s)$ to $q$, the original sampling probabilities are no longer valid. We therefore recompute the trajectory likelihood under the current Policy Model conditioned on $q$ alone:
\begin{equation}
    \log \pi_\theta(\tau \mid q) = \sum_{\ell=1}^{|\tau|} \log \pi_\theta(\tau_\ell \mid q, \tau_{<\ell}).
\end{equation}

We then update $\pi_\theta$ using the verifier reward associated with each pruned trajectory:
\begin{equation}
    \mathcal{L}_{\text{OSI}}(\theta) = -\frac{1}{|\mathcal{B}_t|} \sum_{(q,s,\tau,r) \in \mathcal{B}_t} r \log \pi_\theta(\tau \mid q).
\end{equation}
Since $r \in \{0,1\}$, only verified successful trajectories contribute positive learning signal in practice. This update can be viewed as reward-weighted policy optimization under the \emph{unscaffolded} task condition.

OSI complements VGSL: the latter improves the quality of online skill-guided exploration, while the former turns successful online discoveries into lasting parametric knowledge. Together, they enable continual adaptation under the original task input rather than relying on ever-growing external text memories.

\section{Experiments}
\label{sec:experiments}


\subsection{Experimental Setup}

\textbf{Datasets.} We conduct experiments on LifelongAgentBench \citep{zheng2025lifelongagentbenchevaluatingllmagents}, a benchmark for evaluating lifelong learning in LLM agents. It contains three interactive environments: Database (DB), Operating System (OS), and Knowledge Graph (KG). Each environment requires the agent to solve interdependent tasks under deterministic binary feedback.

\textbf{Baselines.} We compare \texttt{LifeSkill} with two categories of baselines, all built on the same Llama-3.1-8B-Instruct backbone. 
(1) \textbf{Training-free methods (memory retrieval).} These methods improve performance through context augmentation without updating model parameters. We compare against Vanilla (no retrieval), Reasoning Bank \citep{DBLP:journals/corr/abs-2509-25140}, Group Self-Consistency \citep{zheng2025lifelongagentbenchevaluatingllmagents}, AWM \citep{DBLP:conf/icml/WangMFN25}, and Self-Refine \citep{DBLP:conf/nips/MadaanTGHGW0DPY23}. 
(2) \textbf{Training-based methods.} These methods update model parameters but do not perform online dual-model co-evolution. We compare against Rejection Sampling Fine-Tuning (RFT) \citep{DBLP:journals/corr/abs-2308-01825}, SkillRL \citep{xia2026skillrlevolvingagentsrecursive}, and DAPO \citep{DBLP:journals/corr/abs-2503-14476}. Since SkillRL was not originally introduced on LifelongAgentBench, we adapt its recursive skill-augmented training pipeline to this benchmark under the same backbone and evaluation protocol.

\textbf{Implementation Details.} All methods use Llama-3.1-8B-Instruct as the backbone. In \texttt{LifeSkill}, the Policy Model and the Skill Extraction Model are initialized from the same backbone while maintaining separate LoRA adapters with rank 32 and $\alpha=64$. Following the online protocol of LifelongAgentBench, we process tasks sequentially in the benchmark's default order, using temperature 1.0 and top-$p$ 0.9 for rollout generation. We update the Skill Extraction Model online with the standard DAPO token-level clipped objective \citep{DBLP:journals/corr/abs-2503-14476}, using dual clipping with clip ranges $(0.2, 0.28)$. We set the learning rate to $2\times10^{-5}$ and accumulate two samples before each online update. We will release the codes and models on GitHub.

\begin{table}[t]
\vspace{-2mm}
\centering
\small
\setlength{\tabcolsep}{8pt}
\begin{tabular}{@{}l c c c c c@{}}
\toprule
\multirow{2}{*}{\textbf{Method}} & \multirow{2}{*}{\textbf{Experience}} & \multicolumn{4}{c}{\textbf{Accuracy} $\uparrow$} \\
\cmidrule(l){3-6}
 & & \textbf{DB} & \textbf{OS} & \textbf{KG} & \textbf{Avg.} \\
\midrule
\multicolumn{6}{@{}l}{\textit{\small Training-Free Methods}} \\[2pt]
Vanilla                & 0 & 0.16 & 0.44 & 0.27 & 0.29 \\
Self-Refine \citep{DBLP:conf/nips/MadaanTGHGW0DPY23}           & 0 & 0.25 & 0.52 & 0.31 & 0.36 \\
Reasoning Bank \citep{DBLP:journals/corr/abs-2509-25140}        & 4 & 0.25 & 0.45 & 0.30 & 0.33 \\
Group Self-Consistency \citep{zheng2025lifelongagentbenchevaluatingllmagents} & 4 & 0.63 & 0.49 & \textbf{0.36} & 0.49 \\
AWM \citep{DBLP:conf/icml/WangMFN25}                   & 4 & 0.20 & \underline{0.56} & 0.33 & 0.36 \\
\midrule
\multicolumn{6}{@{}l}{\textit{\small Training-Based Methods}} \\[2pt]
RFT \citep{DBLP:journals/corr/abs-2308-01825}                   & 0 & \underline{0.70} & 0.51 & \underline{0.35} & \underline{0.52} \\
DAPO \citep{DBLP:journals/corr/abs-2503-14476}                  & 0 & 0.64 & 0.53 & 0.28 & 0.48 \\
SkillRL \citep{xia2026skillrlevolvingagentsrecursive} & 4 & 0.65 & 0.51 & 0.34 & 0.50 \\
\rowcolor{oursrow}
\texttt{LifeSkill} (Ours) & 0 & \textbf{0.82} & \textbf{0.64} & 0.32 & \textbf{0.59} \\
\bottomrule
\end{tabular}
\vspace{-1mm}
\caption{Main results on LifelongAgentBench across three environments. ``Experience'' denotes the number of past experiences injected into the prompt at inference time. Best results per column are in \textbf{bold}; second-best are \underline{underlined}.}
\label{tab:main_results}
\vspace{-4mm}
\end{table}

\subsection{Main Results}

Table~\ref{tab:main_results} and Figure~\ref{fig:main_results} summarize the main results. \texttt{LifeSkill} achieves the best average accuracy of 0.59, outperforming the strongest training-free baseline by 10 absolute points and the strongest training-based baseline by 7 points.

\textbf{Comparison with training-free methods.}
Training-free methods improve performance by injecting retrieved experiences into the prompt, but they do not update model parameters. Even with up to four retrieved memories, their gains remain limited on challenging environments. On DB, the strongest retrieval-based baseline, Group Self-Consistency, reaches 0.63, which is still 19 points below \texttt{LifeSkill} (0.82). Notably, \texttt{LifeSkill} achieves this result without any retrieved context, suggesting that internalizing skills into model parameters is both more effective and more context-efficient than prompt-based retrieval.

\textbf{Comparison with training-based methods.}
Among training-based methods, RFT achieves 0.70 on DB, while DAPO reaches 0.53 on OS. Both remain substantially below \texttt{LifeSkill}, which attains 0.82 on DB and 0.64 on OS. The gap is especially clear on DB (+0.12 over RFT), where long-horizon tasks require structured reasoning beyond what reward-only trial-and-error optimization can reliably provide. In contrast, the co-evolving Skill Extraction Model supplies more informative guidance for exploration and policy improvement.

\textbf{Generalization across environments.}
\texttt{LifeSkill} ranks first on DB and OS and achieves the best average performance across all three environments. On KG, however, it does not outperform the strongest training-free baseline (0.32 vs.\ 0.36 for Group Self-Consistency). A possible reason is that KG tasks typically involve longer trajectories, which makes the binary reward signal much sparser and weakens the learning signal for both skill extraction and policy internalization. Improving performance in sparse-reward, long-horizon settings remains an important direction for future work.

\subsection{Continual Learning Analysis}

\begin{table}[H]
\vspace{-2mm}
\centering
\footnotesize
\setlength{\tabcolsep}{8pt}
\begin{tabular}{@{}llcc@{}}
\toprule
\textbf{Task} & \textbf{Method} & \textbf{Retention} $\uparrow$ & \textbf{Forgetting} $\downarrow$ \\
\midrule
\multirow{3}{*}{DB}
& RFT & 0.552 & 0.156 \\
& DAPO & 0.604 & \textbf{0.042} \\
& \texttt{LifeSkill} & \textbf{0.708} & 0.104 \\
\midrule
\multirow{3}{*}{OS}
& RFT & 0.406 & 0.104 \\
& DAPO & 0.469 & \textbf{0.042} \\
& \texttt{LifeSkill} & \textbf{0.552} & 0.083 \\
\midrule
\multirow{3}{*}{KG}
& RFT & \textbf{0.293} & 0.076 \\
& DAPO & 0.250 & \textbf{0.043} \\
& \texttt{LifeSkill} & 0.283 & 0.065 \\
\bottomrule
\end{tabular}
\vspace{-1mm}
\caption{Post-update retention and forgetting. Retention and forgetting are evaluated on previous-window tasks after the full online stream has been processed.}
\label{tab:retention_forgetting}
\vspace{-4mm}
\end{table}

Table~\ref{tab:retention_forgetting} provides a post-update retention analysis. After processing the full online stream, we re-evaluate tasks from previous windows and measure both the final retained performance and the corresponding forgetting. On DB and OS, \texttt{LifeSkill} achieves the highest Past Retention, suggesting that skill-guided exploration followed by unscaffolded internalization improves retained earlier-window capability. DAPO obtains lower Forgetting, while \texttt{LifeSkill} reaches higher retained accuracy, reflecting a stability-adaptation trade-off between conservative updates and stronger online improvement. On KG, \texttt{LifeSkill} remains below RFT in retention, which is consistent with the sparse-reward limitation discussed above.

\subsection{Ablation Studies}
\label{sec:analysis}

Table~\ref{tab:ablation_vertical_detailed} shows that each component of \texttt{LifeSkill} contributes to the final performance. Replacing Verifier-Guided Skill Learning (VGSL) with LLM-as-Judge supervision reduces accuracy from 0.82/0.64 to 0.70/0.62 on DB/OS, indicating that execution-grounded verifier rewards provide a more reliable signal for skill extraction than subjective scoring. Removing the Skill Extraction Model leads to the largest degradation, dropping performance to 0.64 on DB and 0.53 on OS, which highlights the importance of dual-model co-evolution for discovering useful guidance beyond reward-only policy optimization. Disabling Online Skill Internalization (OSI) also hurts performance, reducing accuracy to 0.79 on DB and 0.60 on OS, showing that skill extraction alone is not sufficient unless the discovered behaviors are further absorbed into the policy parameters. 

\begin{table}[H]
\vspace{-1mm}
\centering
\small
\setlength{\tabcolsep}{8pt}
\begin{tabular}{@{}llcc@{}}
\toprule
\textbf{Component} & \textbf{Configuration} & \textbf{DB} & \textbf{OS} \\
\midrule
Reward Signal         & w/o VGSL (LLM-Judge) & 0.70 & 0.62 \\
Architecture          & w/o Skill Extraction & 0.64 & 0.53 \\
Skill Internalization & w/o OSI             & 0.79 & 0.60 \\
\midrule
\textbf{\texttt{LifeSkill} (Ours)} & Full Model & \textbf{0.82} & \textbf{0.64} \\
\bottomrule
\end{tabular}
\vspace{-1mm}
\caption{Results of ablation studies.}
\label{tab:ablation_vertical_detailed}
\vspace{-3mm}
\end{table}

\begin{table}[H]
\vspace{-2mm}
\centering
\small
\setlength{\tabcolsep}{12pt}
\begin{tabular}{@{}lccc@{}}
\toprule
\textbf{Method} & \textbf{DB Avg. $\pm$ Std.} & \textbf{OS Avg. $\pm$ Std.} & \textbf{Avg.} \\
\midrule
DAPO & $0.640{\pm}0.061$ & $0.530{\pm}0.050$ & 0.585 \\
\rowcolor{oursrow}
\texttt{LifeSkill} & $\textbf{0.820}{\pm}0.044$ & $\textbf{0.640}{\pm}0.056$ & \textbf{0.730} \\
\bottomrule
\end{tabular}
\vspace{-1mm}
\caption{Task-order sensitivity on DB and OS. We report the mean accuracy and sample standard deviation across three different task orders.}
\label{tab:task_order_sensitivity}
\vspace{-4mm}
\end{table}

\begin{table}[H]
\vspace{-2mm}
\centering
\footnotesize
\setlength{\tabcolsep}{5pt}
\begin{tabular}{@{}lcccccccc@{}}
\toprule
\multirow{2}{*}{\textbf{Method}} &
\multicolumn{4}{c}{\textbf{Llama-3.1-8B-Instruct}} &
\multicolumn{4}{c}{\textbf{Qwen2.5-7B-Instruct}} \\
\cmidrule(lr){2-5}\cmidrule(l){6-9}
& \textbf{DB} & \textbf{OS} & \textbf{KG} & \textbf{Avg.}
& \textbf{DB} & \textbf{OS} & \textbf{KG} & \textbf{Avg.} \\
\midrule
Vanilla & 0.16 & 0.44 & 0.27 & 0.29 & 0.74 & 0.50 & 0.22 & 0.49 \\
RFT & 0.70 & 0.51 & \textbf{0.35} & 0.52 & 0.77 & 0.51 & 0.26 & 0.51 \\
DAPO & 0.64 & 0.53 & 0.28 & 0.48 & 0.76 & 0.53 & 0.31 & 0.53 \\
\rowcolor{oursrow}
\texttt{LifeSkill} & \textbf{0.82} & \textbf{0.64} & 0.32 & \textbf{0.59} & \textbf{0.83} & \textbf{0.65} & \textbf{0.38} & \textbf{0.62} \\
\bottomrule
\end{tabular}
\vspace{-1mm}
\caption{Backbone ablation results. Qwen2.5-7B-Instruct follows the same online evaluation protocol as the Llama-3.1-8B-Instruct experiments.}
\label{tab:backbone_ablation}
\vspace{-4mm}
\end{table}

We further analyze sensitivity to the online task order in Table~\ref{tab:task_order_sensitivity}. \texttt{LifeSkill} maintains higher mean accuracy than DAPO on both DB and OS, improving the average of the two environments from 0.585 to 0.730. The standard deviations are comparable across methods, indicating that the relative gain is stable under different stream permutations.

We further conduct a backbone ablation in Table~\ref{tab:backbone_ablation}, comparing Llama-3.1-8B-Instruct with Qwen2.5-7B-Instruct under the same online protocol for Vanilla, RFT, DAPO, and \texttt{LifeSkill}. The Qwen backbone yields much stronger Vanilla performance on DB, a smaller improvement on OS, and lower Vanilla performance on KG, indicating that the effect of backbone choice varies across environments. For \texttt{LifeSkill}, switching from Llama to Qwen improves accuracy on all three environments, with the largest gain appearing on KG.


\subsection{Hyperparameter Analysis}

\begin{figure}[H]
    \centering
    \begin{subfigure}[t]{0.40\textwidth}
        \centering
        \includegraphics[width=\linewidth]{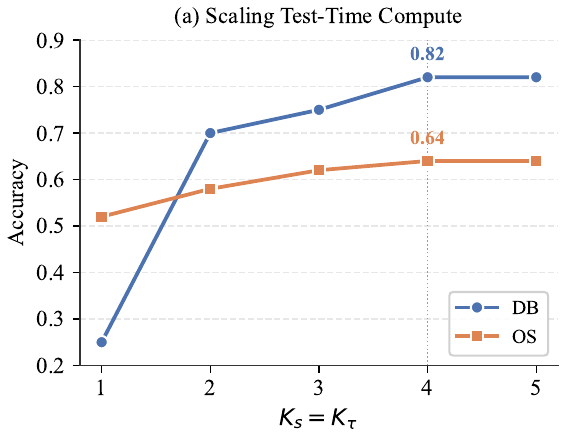}
        \vspace{-4mm}
        \caption{Sensitivity of $K_s$ and $K_\tau$.}
        \label{fig:k_scaling}
    \end{subfigure}
    \begin{subfigure}[t]{0.40\textwidth}
        \centering
        \includegraphics[width=\linewidth]{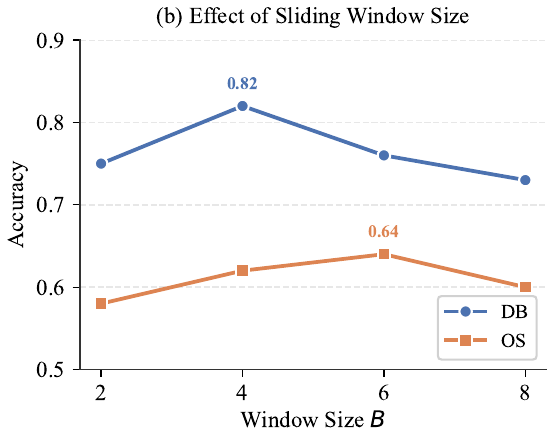}
        \vspace{-4mm}
        \caption{Sensitivity of the window size $B$.}
        \label{fig:window_size}
    \end{subfigure}
    \vspace{-2mm}
    \caption{Hyperparameter analysis of \texttt{LifeSkill}.}
    \label{fig:hyperparameter_analysis}
    \vspace{-5mm}
\end{figure}


\textbf{Scaling Test-Time Compute.}
We study the effect of test-time compute by varying the number of sampled candidate skills $K_s$ and skill-conditioned trajectories $K_\tau$. For simplicity, we set $K_s = K_\tau$ and vary them jointly from 1 to 5. As shown in Figure~\ref{fig:k_scaling}, increasing the rollout budget consistently improves performance on both DB and OS. On DB, accuracy rises markedly from 0.25 at $K_s{=}K_\tau{=}1$ to 0.82 at 4, and then plateaus at 5. OS shows a similar but milder trend, reaching 0.64 at 4 and remaining stable thereafter. These results suggest that a larger exploration budget generally leads to better performance, while the limited gains beyond 4 indicate a practical trade-off between compute cost and accuracy.


\textbf{Effect of Window Size.}
We further analyze the effect of the window size $B$, which controls the trade-off between update frequency and optimization stability. Smaller windows enable more frequent online updates but may introduce noisier gradients, whereas larger windows provide more stable updates at the cost of slower adaptation. As shown in Figure~\ref{fig:window_size}, DB achieves its best performance at $B{=}4$ (0.82) and gradually declines as the window becomes larger, while OS peaks at $B{=}6$ (0.64). This difference suggests that the two environments favor different adaptation granularities: DB appears to benefit more from faster updates, whereas OS benefits from slightly larger windows.

\subsection{Qualitative Analysis}

We use two case studies to understand what the Skill Extraction Model learns and how the extracted skills affect downstream execution. Table~\ref{tab:skill_evolution} traces representative skills from different training stages. Table~\ref{tab:skill_correction} shows a concrete failure-recovery example in which the extracted skill changes the policy's SQL decision and corrects the final answer.

\begin{table}[t]
\vspace{-1mm}
\centering
\footnotesize
\renewcommand{\arraystretch}{1.}
\setlength{\tabcolsep}{6pt}
\begin{tabular}{@{}p{0.08\textwidth} p{0.89\textwidth}@{}}
\toprule
\textbf{Stage} & \textbf{Extracted Skill} \\
\midrule
Early & Apply filtering in a subquery with \texttt{WHERE} before ordering and limiting the results. \\
\midrule
Middle & Use \texttt{HAVING}, rather than \texttt{WHERE}, to filter aggregate results after grouping. \\
\midrule
Late & Map strict count constraints such as `exceeds' and `fewer than' to strict inequalities in \texttt{HAVING}, rather than inclusive \texttt{BETWEEN} conditions. \\
\bottomrule
\end{tabular}
\vspace{-1mm}
\caption{Representative extracted skills from early, middle, and late training stages.}
\label{tab:skill_evolution}
\vspace{-1mm}
\end{table}

\begin{table}[t]
\centering
\footnotesize
\renewcommand{\arraystretch}{1.}
\setlength{\tabcolsep}{6pt}
\begin{tabular}{@{}p{0.07\textwidth} p{0.89\textwidth}@{}}
\toprule
\textbf{Item} & \textbf{Content} \\
\midrule
Query & For \texttt{Fiction} books published after 2000, return yearly counts where the total exceeds 5 but is fewer than 10. \\
\midrule
Initial SQL &
\parbox[t]{\linewidth}{
\texttt{SELECT genre, published\_year, COUNT(*) AS total\_count FROM}\\
\texttt{library\_books WHERE genre = 'Fiction' AND published\_year > 2000}\\
\texttt{GROUP BY genre, published\_year HAVING COUNT(*) BETWEEN 5 AND 9;}
} \\
\midrule
Extracted skill & Map strict count constraints such as `exceeds' and `fewer than' to strict inequalities in \texttt{HAVING}. \\
\midrule
Corrected SQL &
\parbox[t]{\linewidth}{
\texttt{SELECT genre, published\_year, COUNT(*) AS total\_count FROM}\\
\texttt{library\_books WHERE genre = 'Fiction' AND published\_year > 2000}\\
\texttt{GROUP BY genre, published\_year HAVING COUNT(*) > 5 AND COUNT(*) < 10;}
} \\
\bottomrule
\end{tabular}
\vspace{-1mm}
\caption{A skill-guided correction example on the DB environment of LifelongAgentBench. The extracted skill fixes the semantic interpretation of a strict count constraint.}
\label{tab:skill_correction}
\vspace{-3mm}
\end{table}

Table~\ref{tab:skill_evolution} shows that the extracted skills become progressively more precise and execution-relevant over training. Early-stage skills are relatively generic and can be partially misleading, while middle-stage skills begin to capture operator-level distinctions such as the different roles of \texttt{HAVING} and \texttt{WHERE}. By the late stage, the extracted skill aligns more closely with the underlying task semantics, correctly mapping strict natural-language constraints to strict inequalities. This trend suggests that verifier-guided training gradually suppresses generic reflections and encourages compact skills that are more useful for downstream execution.

Table~\ref{tab:skill_correction} further shows that the extracted skill is not merely descriptive. The initial policy applies an inclusive interpretation to a strict cardinality constraint, which leads to an incorrect answer. After injecting the extracted skill, the retry adopts the correct semantic reading and solves the task successfully. This example indicates that the extracted skill operates at the level of constraint interpretation, steering the policy toward a corrected execution rather than merely paraphrasing the previous failure.

\section{Conclusions and Further Work}

In this paper, we proposed \texttt{LifeSkill}, a skill-enhanced test-time co-evolution framework for online lifelong LLM agents. By combining Verifier-Guided Skill Learning and Online Skill Internalization, \texttt{LifeSkill} enables the agent to extract useful skills from failed interactions and absorb successful skill-guided behaviors into its policy parameters during deployment. Experiments on LifelongAgentBench show that \texttt{LifeSkill} consistently outperforms both memory-based and RL-based baselines, demonstrating the effectiveness of coupling execution-grounded skill extraction with online parametric adaptation.
For future work, it would be valuable to improve performance in sparse-reward long-horizon settings and to reduce the test-time compute cost of skill-guided exploration. 

\bibliographystyle{colm2026_conference}
\bibliography{colm2026_conference}

\begin{thebibliography}{47}
\providecommand{\natexlab}[1]{#1}
\providecommand{\url}[1]{\texttt{#1}}
\expandafter\ifx\csname urlstyle\endcsname\relax
  \providecommand{\doi}[1]{doi: #1}\else
  \providecommand{\doi}{doi: \begingroup \urlstyle{rm}\Url}\fi

\bibitem[Aljundi et~al.(2019)Aljundi, Caccia, Belilovsky, Caccia, Lin, Charlin, and Tuytelaars]{aljundi2019mir}
Rahaf Aljundi, Lucas Caccia, Eugene Belilovsky, Massimo Caccia, Min Lin, Laurent Charlin, and Tinne Tuytelaars.
\newblock \emph{Online continual learning with maximally interfered retrieval}.
\newblock Curran Associates Inc., Red Hook, NY, USA, 2019.

\bibitem[Besta et~al.(2024)Besta, Blach, Kubicek, Gerstenberger, Podstawski, Gianinazzi, et~al.]{besta2024graph}
Maciej Besta, Nils Blach, Ales Kubicek, Robert Gerstenberger, Michal Podstawski, Lukas Gianinazzi, et~al.
\newblock Graph of thoughts: Solving elaborate problems with large language models.
\newblock \emph{Proceedings of the AAAI Conference on Artificial Intelligence}, 2024.

\bibitem[Cai et~al.(2025)Cai, Hao, Zhou, Yan, Lei, Zhen, Han, Yang, Li, Pan, et~al.]{cai2025building}
Yuxuan Cai, Yipeng Hao, Jie Zhou, Hang Yan, Zhikai Lei, Rui Zhen, Zhenhua Han, Yutao Yang, Junsong Li, Qianjun Pan, et~al.
\newblock Building self-evolving agents via experience-driven lifelong learning: A framework and benchmark.
\newblock \emph{arXiv preprint arXiv:2508.19005}, 2025.

\bibitem[Chen et~al.(2023)Chen, Ma, Wang, and Cohen]{chen2023pot}
Wenhu Chen, Xueguang Ma, Xinyi Wang, and William~W. Cohen.
\newblock Program of thoughts prompting: Disentangling computation from reasoning for numerical reasoning tasks.
\newblock \emph{Transactions on Machine Learning Research}, 2023.

\bibitem[DeepSeek{-}AI(2025)]{DBLP:journals/corr/abs-2501-12948}
DeepSeek{-}AI.
\newblock Deepseek-r1: Incentivizing reasoning capability in llms via reinforcement learning.
\newblock \emph{CoRR}, abs/2501.12948, 2025.
\newblock \doi{10.48550/ARXIV.2501.12948}.
\newblock URL \url{https://doi.org/10.48550/arXiv.2501.12948}.

\bibitem[Fang et~al.(2025)Fang, Peng, Zhang, Wang, Yi, Zhang, Xu, Wu, Liu, Li, et~al.]{fang2025comprehensive}
Jinyuan Fang, Yanwen Peng, Xi~Zhang, Yingxu Wang, Xinhao Yi, Guibin Zhang, Yi~Xu, Bin Wu, Siwei Liu, Zihao Li, et~al.
\newblock A comprehensive survey of self-evolving ai agents: A new paradigm bridging foundation models and lifelong agentic systems.
\newblock \emph{arXiv preprint arXiv:2508.07407}, 2025.

\bibitem[Gao et~al.(2025)Gao, Geng, Hua, Hu, Juan, Liu, Liu, Qiu, Qi, Wu, et~al.]{gao2025survey}
Huan-ang Gao, Jiayi Geng, Wenyue Hua, Mengkang Hu, Xinzhe Juan, Hongzhang Liu, Shilong Liu, Jiahao Qiu, Xuan Qi, Yiran Wu, et~al.
\newblock A survey of self-evolving agents: What, when, how, and where to evolve on the path to artificial super intelligence.
\newblock \emph{arXiv preprint arXiv:2507.21046}, 1, 2025.

\bibitem[Gao et~al.(2023)Gao, Madaan, Zhou, Alon, Liu, Yang, Callan, and Neubig]{gao2023pal}
Luyu Gao, Aman Madaan, Shuyan Zhou, Uri Alon, Pengfei Liu, Yiming Yang, Jamie Callan, and Graham Neubig.
\newblock {PAL}: Program-aided language models.
\newblock In \emph{International Conference on Machine Learning}, 2023.

\bibitem[Kirkpatrick et~al.(2017)Kirkpatrick, Pascanu, Rabinowitz, Veness, Desjardins, et~al.]{kirkpatrick2017ewc}
James Kirkpatrick, Razvan Pascanu, Neil Rabinowitz, Joel Veness, Guillaume Desjardins, et~al.
\newblock Overcoming catastrophic forgetting in neural networks.
\newblock \emph{Proceedings of the National Academy of Sciences}, 114\penalty0 (13):\penalty0 3521--3526, 2017.

\bibitem[Koh et~al.(2024)Koh, Lo, Jang, Duvvur, Lim, Huang, et~al.]{koh2024visualwebarena}
Jing~Yu Koh, Robert Lo, Lawrence Jang, Vikram Duvvur, Ming~Chong Lim, Po-Yu Huang, et~al.
\newblock Visualwebarena: Evaluating multimodal agents on realistic visual web tasks.
\newblock \emph{arXiv preprint arXiv:2401.13649}, 2024.

\bibitem[Li et~al.(2025)Li, Song, Wang, Niu, Chen, Yang, et~al.]{li2025memos}
Zhiyu Li, Shichao Song, Hanyu Wang, Simin Niu, Ding Chen, Jiawei Yang, et~al.
\newblock Memos: An operating system for memory-augmented generation in large language models.
\newblock \emph{arXiv preprint arXiv:2505.22101}, 2025.

\bibitem[Liu et~al.(2024)Liu, Yu, Zhang, Xu, Lei, Lai, et~al.]{liu2023agentbench}
Xiao Liu, Hao Yu, Hanchen Zhang, Yifan Xu, Xuanyu Lei, Hanyu Lai, et~al.
\newblock Agentbench: Evaluating {LLM}s as agents.
\newblock In \emph{International Conference on Learning Representations}, 2024.

\bibitem[Liu et~al.(2025)Liu, Si, Narasimhan, and Yao]{DBLP:conf/acl/LiuSNY25}
Yitao Liu, Chenglei Si, Karthik~R. Narasimhan, and Shunyu Yao.
\newblock Contextual experience replay for self-improvement of language agents.
\newblock In Wanxiang Che, Joyce Nabende, Ekaterina Shutova, and Mohammad~Taher Pilehvar (eds.), \emph{Proceedings of the 63rd Annual Meeting of the Association for Computational Linguistics (Volume 1: Long Papers), {ACL} 2025, Vienna, Austria, July 27 - August 1, 2025}, pp.\  14179--14198. Association for Computational Linguistics, 2025.
\newblock URL \url{https://aclanthology.org/2025.acl-long.694/}.

\bibitem[Madaan et~al.(2023)Madaan, Tandon, Gupta, Hallinan, Gao, Wiegreffe, Alon, Dziri, Prabhumoye, Yang, Gupta, Majumder, Hermann, Welleck, Yazdanbakhsh, and Clark]{DBLP:conf/nips/MadaanTGHGW0DPY23}
Aman Madaan, Niket Tandon, Prakhar Gupta, Skyler Hallinan, Luyu Gao, Sarah Wiegreffe, Uri Alon, Nouha Dziri, Shrimai Prabhumoye, Yiming Yang, Shashank Gupta, Bodhisattwa~Prasad Majumder, Katherine Hermann, Sean Welleck, Amir Yazdanbakhsh, and Peter Clark.
\newblock Self-refine: Iterative refinement with self-feedback.
\newblock In Alice Oh, Tristan Naumann, Amir Globerson, Kate Saenko, Moritz Hardt, and Sergey Levine (eds.), \emph{Advances in Neural Information Processing Systems 36: Annual Conference on Neural Information Processing Systems 2023, NeurIPS 2023, New Orleans, LA, USA, December 10 - 16, 2023}, 2023.
\newblock URL \url{http://papers.nips.cc/paper\_files/paper/2023/hash/91edff07232fb1b55a505a9e9f6c0ff3-Abstract-Conference.html}.

\bibitem[Ouyang et~al.(2022)Ouyang, Wu, Jiang, Almeida, Wainwright, Mishkin, et~al.]{ouyang2022instructgpt}
Long Ouyang, Jeff Wu, Xu~Jiang, Diogo Almeida, Carroll~L. Wainwright, Pamela Mishkin, et~al.
\newblock Training language models to follow instructions with human feedback.
\newblock In \emph{Advances in Neural Information Processing Systems}, 2022.

\bibitem[Ouyang et~al.(2025)Ouyang, Yan, Hsu, Chen, Jiang, Wang, Han, Le, Daruki, Tang, Tirumalashetty, Lee, Rofouei, Lin, Han, Lee, and Pfister]{DBLP:journals/corr/abs-2509-25140}
Siru Ouyang, Jun Yan, I{-}Hung Hsu, Yanfei Chen, Ke~Jiang, Zifeng Wang, Rujun Han, Long~T. Le, Samira Daruki, Xiangru Tang, Vishy Tirumalashetty, George Lee, Mahsan Rofouei, Hangfei Lin, Jiawei Han, Chen{-}Yu Lee, and Tomas Pfister.
\newblock Reasoningbank: Scaling agent self-evolving with reasoning memory.
\newblock \emph{CoRR}, abs/2509.25140, 2025.
\newblock \doi{10.48550/ARXIV.2509.25140}.
\newblock URL \url{https://doi.org/10.48550/arXiv.2509.25140}.

\bibitem[Packer et~al.(2024)Packer, Wooders, Lin, Fang, Patil, Stoica, and Gonzalez]{packer2024memgpt}
Charles Packer, Sarah Wooders, Kevin Lin, Vivian Fang, Shishir~G. Patil, Ion Stoica, and Joseph~E. Gonzalez.
\newblock Memgpt: Towards {LLM}s as operating systems.
\newblock \emph{arXiv preprint arXiv:2310.08560}, 2024.

\bibitem[Park et~al.(2023)Park, O'Brien, Cai, Morris, Liang, and Bernstein]{park2023generative}
Joon~Sung Park, Joseph~C. O'Brien, Carrie~J. Cai, Meredith~Ringel Morris, Percy Liang, and Michael~S. Bernstein.
\newblock Generative agents: Interactive simulacra of human behavior.
\newblock In \emph{Proceedings of the 36th Annual ACM Symposium on User Interface Software and Technology}, 2023.

\bibitem[Rolnick et~al.(2019)Rolnick, Ahuja, Schwarz, Lillicrap, and Wayne]{rolnick2019experience}
David Rolnick, Arun Ahuja, Jonathan Schwarz, Timothy~P. Lillicrap, and Greg Wayne.
\newblock Experience replay for continual learning.
\newblock In \emph{Advances in Neural Information Processing Systems}, 2019.

\bibitem[Schick et~al.(2023)Schick, Dwivedi-Yu, Dess{\`i}, Raileanu, Lomeli, Hambro, et~al.]{schick2023toolformer}
Timo Schick, Jane Dwivedi-Yu, Roberto Dess{\`i}, Roberta Raileanu, Maria Lomeli, Eric Hambro, et~al.
\newblock Toolformer: Language models can teach themselves to use tools.
\newblock In \emph{Advances in Neural Information Processing Systems}, 2023.

\bibitem[Schulman et~al.(2017)Schulman, Wolski, Dhariwal, Radford, and Klimov]{schulman2017ppo}
John Schulman, Filip Wolski, Prafulla Dhariwal, Alec Radford, and Oleg Klimov.
\newblock Proximal policy optimization algorithms.
\newblock \emph{arXiv preprint arXiv:1707.06347}, 2017.

\bibitem[Schwarz et~al.(2018)Schwarz, Czarnecki, Luketina, Grabska-Barwinska, Teh, Pascanu, and Hadsell]{schwarz2018progresscompress}
Jonathan Schwarz, Wojciech~M. Czarnecki, Jelena Luketina, Agnieszka Grabska-Barwinska, Yee~Whye Teh, Razvan Pascanu, and Raia Hadsell.
\newblock Progress \& compress: A scalable framework for continual learning.
\newblock In \emph{International Conference on Machine Learning}, 2018.

\bibitem[Shinn et~al.(2023)Shinn, Cassano, Gopinath, Narasimhan, and Yao]{DBLP:conf/nips/ShinnCGNY23}
Noah Shinn, Federico Cassano, Ashwin Gopinath, Karthik Narasimhan, and Shunyu Yao.
\newblock Reflexion: language agents with verbal reinforcement learning.
\newblock In Alice Oh, Tristan Naumann, Amir Globerson, Kate Saenko, Moritz Hardt, and Sergey Levine (eds.), \emph{Advances in Neural Information Processing Systems 36: Annual Conference on Neural Information Processing Systems 2023, NeurIPS 2023, New Orleans, LA, USA, December 10 - 16, 2023}, 2023.
\newblock URL \url{http://papers.nips.cc/paper\_files/paper/2023/hash/1b44b878bb782e6954cd888628510e90-Abstract-Conference.html}.

\bibitem[Sun et~al.(2020)Sun, Wang, Liu, Miller, Efros, and Hardt]{sun2020ttt}
Yu~Sun, Xiaolong Wang, Zhuang Liu, John Miller, Alexei~A. Efros, and Moritz Hardt.
\newblock Test-time training with self-supervision for generalization under distribution shifts.
\newblock In \emph{International Conference on Machine Learning}, 2020.

\bibitem[Sutton et~al.(1999)Sutton, Precup, and Singh]{sutton1999options}
Richard~S. Sutton, Doina Precup, and Satinder Singh.
\newblock Between {MDP}s and semi-{MDP}s: A framework for temporal abstraction in reinforcement learning.
\newblock \emph{Artificial Intelligence}, 112\penalty0 (1--2):\penalty0 181--211, 1999.

\bibitem[Suzgun et~al.(2025)Suzgun, Yuksekgonul, Bianchi, Jurafsky, and Zou]{suzgun2025dynamic}
Mirac Suzgun, Mert Yuksekgonul, Federico Bianchi, Dan Jurafsky, and James Zou.
\newblock Dynamic cheatsheet: Test-time learning with adaptive memory.
\newblock \emph{arXiv preprint arXiv:2504.07952}, 2025.

\bibitem[Tang et~al.(2025)Tang, Qin, Peng, Zhou, Shao, Du, Wei, Xia, Wu, Zhu, Zhang, Liu, Wang, Hong, Wu, Cheng, Wang, and Zhou]{DBLP:journals/corr/abs-2507-06229}
Xiangru Tang, Tianrui Qin, Tianhao Peng, Ziyang Zhou, Daniel Shao, Tingting Du, Xinming Wei, Peng Xia, Fang Wu, He~Zhu, Ge~Zhang, Jiaheng Liu, Xingyao Wang, Sirui Hong, Chenglin Wu, Hao Cheng, Chi Wang, and Wangchunshu Zhou.
\newblock Agent {KB:} leveraging cross-domain experience for agentic problem solving.
\newblock \emph{CoRR}, abs/2507.06229, 2025.
\newblock \doi{10.48550/ARXIV.2507.06229}.
\newblock URL \url{https://doi.org/10.48550/arXiv.2507.06229}.

\bibitem[Wang et~al.(2021)Wang, Shelhamer, Liu, Olshausen, and Darrell]{wang2021tent}
Dequan Wang, Evan Shelhamer, Shaoteng Liu, Bruno~A. Olshausen, and Trevor Darrell.
\newblock Tent: Fully test-time adaptation by entropy minimization.
\newblock In \emph{International Conference on Learning Representations}, 2021.

\bibitem[Wang et~al.(2023{\natexlab{a}})Wang, Xie, Jiang, Mandlekar, Xiao, Zhu, Fan, and Anandkumar]{wang2023voyager}
Guanzhi Wang, Yuqi Xie, Yunfan Jiang, Ajay Mandlekar, Chaowei Xiao, Yuke Zhu, Linxi Fan, and Anima Anandkumar.
\newblock Voyager: An open-ended embodied agent with large language models.
\newblock \emph{arXiv preprint arXiv:2305.16291}, 2023{\natexlab{a}}.

\bibitem[Wang et~al.(2024)Wang, Ma, Feng, Zhang, Yang, Zhang, et~al.]{wang2024survey_agents}
Lei Wang, Chengbang Ma, Xueyang Feng, Zeyu Zhang, Hao Yang, Jingsen Zhang, et~al.
\newblock A survey on large language model based autonomous agents.
\newblock \emph{Frontiers of Computer Science}, 2024.

\bibitem[Wang et~al.(2023{\natexlab{b}})Wang, Wei, Schuurmans, Le, Chi, Narang, et~al.]{wang2023selfconsistency}
Xuezhi Wang, Jason Wei, Dale Schuurmans, Quoc Le, Ed~Chi, Sharan Narang, et~al.
\newblock Self-consistency improves chain of thought reasoning in language models.
\newblock In \emph{International Conference on Learning Representations}, 2023{\natexlab{b}}.

\bibitem[Wang et~al.(2025)Wang, Mao, Fried, and Neubig]{DBLP:conf/icml/WangMFN25}
Zora~Zhiruo Wang, Jiayuan Mao, Daniel Fried, and Graham Neubig.
\newblock Agent workflow memory.
\newblock In Aarti Singh, Maryam Fazel, Daniel Hsu, Simon Lacoste{-}Julien, Felix Berkenkamp, Tegan Maharaj, Kiri Wagstaff, and Jerry Zhu (eds.), \emph{Forty-second International Conference on Machine Learning, {ICML} 2025, Vancouver, BC, Canada, July 13-19, 2025}, volume 267 of \emph{Proceedings of Machine Learning Research}. {PMLR} / OpenReview.net, 2025.
\newblock URL \url{https://proceedings.mlr.press/v267/wang25bx.html}.

\bibitem[Wei et~al.(2022)Wei, Wang, Schuurmans, Bosma, Ichter, Xia, et~al.]{wei2022cot}
Jason Wei, Xuezhi Wang, Dale Schuurmans, Maarten Bosma, Brian Ichter, Fei Xia, et~al.
\newblock Chain-of-thought prompting elicits reasoning in large language models.
\newblock In \emph{Advances in Neural Information Processing Systems}, 2022.

\bibitem[Xia et~al.(2026)Xia, Chen, Wang, Liu, Zeng, Wang, Han, Zhou, Zhao, Chen, Zheng, Xie, and Yao]{xia2026skillrlevolvingagentsrecursive}
Peng Xia, Jianwen Chen, Hanyang Wang, Jiaqi Liu, Kaide Zeng, Yu~Wang, Siwei Han, Yiyang Zhou, Xujiang Zhao, Haifeng Chen, Zeyu Zheng, Cihang Xie, and Huaxiu Yao.
\newblock Skillrl: Evolving agents via recursive skill-augmented reinforcement learning, 2026.
\newblock URL \url{https://arxiv.org/abs/2602.08234}.

\bibitem[Yang et~al.(2025)Yang, Zhou, Ding, Huai, Liu, Chen, Xie, and He]{yang2025recent}
Yutao Yang, Jie Zhou, Xuanwen Ding, Tianyu Huai, Shunyu Liu, Qin Chen, Yuan Xie, and Liang He.
\newblock Recent advances of foundation language models-based continual learning: A survey.
\newblock \emph{ACM Computing Surveys}, 57\penalty0 (5):\penalty0 1--38, 2025.

\bibitem[Yao et~al.(2023{\natexlab{a}})Yao, Yu, Zhao, Shafran, Griffiths, Cao, and Narasimhan]{yao2023tree}
Shunyu Yao, Dian Yu, Jeffrey Zhao, Izhak Shafran, Thomas~L. Griffiths, Yuan Cao, and Karthik Narasimhan.
\newblock Tree of thoughts: Deliberate problem solving with large language models.
\newblock \emph{arXiv preprint arXiv:2305.10601}, 2023{\natexlab{a}}.

\bibitem[Yao et~al.(2023{\natexlab{b}})Yao, Zhao, Yu, Du, Shafran, Narasimhan, and Cao]{yao2023react}
Shunyu Yao, Jeffrey Zhao, Dian Yu, Nan Du, Izhak Shafran, Karthik Narasimhan, and Yuan Cao.
\newblock {ReAct}: Synergizing reasoning and acting in language models.
\newblock In \emph{International Conference on Learning Representations}, 2023{\natexlab{b}}.

\bibitem[Yu et~al.(2025)Yu, Zhang, Zhu, Yuan, Zuo, Yue, Fan, Liu, Liu, Liu, Lin, Lin, Ma, Sheng, Tong, Zhang, Zhang, Zhang, Zhu, Zhu, Chen, Chen, Wang, Yu, Dai, Song, Wei, Zhou, Liu, Ma, Zhang, Yan, Qiao, Wu, and Wang]{DBLP:journals/corr/abs-2503-14476}
Qiying Yu, Zheng Zhang, Ruofei Zhu, Yufeng Yuan, Xiaochen Zuo, Yu~Yue, Tiantian Fan, Gaohong Liu, Lingjun Liu, Xin Liu, Haibin Lin, Zhiqi Lin, Bole Ma, Guangming Sheng, Yuxuan Tong, Chi Zhang, Mofan Zhang, Wang Zhang, Hang Zhu, Jinhua Zhu, Jiaze Chen, Jiangjie Chen, Chengyi Wang, Hongli Yu, Weinan Dai, Yuxuan Song, Xiangpeng Wei, Hao Zhou, Jingjing Liu, Wei{-}Ying Ma, Ya{-}Qin Zhang, Lin Yan, Mu~Qiao, Yonghui Wu, and Mingxuan Wang.
\newblock {DAPO:} an open-source {LLM} reinforcement learning system at scale.
\newblock \emph{CoRR}, abs/2503.14476, 2025.
\newblock \doi{10.48550/ARXIV.2503.14476}.
\newblock URL \url{https://doi.org/10.48550/arXiv.2503.14476}.

\bibitem[Yuan et~al.(2023)Yuan, Yuan, Li, Dong, Tan, and Zhou]{DBLP:journals/corr/abs-2308-01825}
Zheng Yuan, Hongyi Yuan, Chengpeng Li, Guanting Dong, Chuanqi Tan, and Chang Zhou.
\newblock Scaling relationship on learning mathematical reasoning with large language models.
\newblock \emph{CoRR}, abs/2308.01825, 2023.
\newblock \doi{10.48550/ARXIV.2308.01825}.
\newblock URL \url{https://doi.org/10.48550/arXiv.2308.01825}.

\bibitem[Zelikman et~al.(2022)Zelikman, Wu, Mu, and Goodman]{zelikman2022star}
Eric Zelikman, Yuhuai Wu, Jesse Mu, and Noah Goodman.
\newblock Star: Bootstrapping reasoning with reasoning.
\newblock \emph{Advances in Neural Information Processing Systems}, 35:\penalty0 15476--15488, 2022.

\bibitem[Zhao et~al.(2024)Zhao, Huang, Xu, Lin, Liu, and Huang]{zhao2024expel}
Andrew Zhao, Daniel Huang, Quentin Xu, Matthieu Lin, Yong-Jin Liu, and Gao Huang.
\newblock Expel: {LLM} agents are experiential learners.
\newblock In \emph{Proceedings of the AAAI Conference on Artificial Intelligence}, 2024.

\bibitem[Zheng et~al.(2025)Zheng, Cai, Li, Zhang, Li, Zhang, Song, and Ma]{zheng2025lifelongagentbenchevaluatingllmagents}
Junhao Zheng, Xidi Cai, Qiuke Li, Duzhen Zhang, ZhongZhi Li, Yingying Zhang, Le~Song, and Qianli Ma.
\newblock Lifelongagentbench: Evaluating llm agents as lifelong learners, 2025.
\newblock URL \url{https://arxiv.org/abs/2505.11942}.

\bibitem[Zheng et~al.(2023)Zheng, Chiang, Sheng, Zhuang, Wu, Zhuang, Lin, Li, Li, Xing, Zhang, Gonzalez, and Stoica]{DBLP:conf/nips/ZhengC00WZL0LXZ23}
Lianmin Zheng, Wei{-}Lin Chiang, Ying Sheng, Siyuan Zhuang, Zhanghao Wu, Yonghao Zhuang, Zi~Lin, Zhuohan Li, Dacheng Li, Eric~P. Xing, Hao Zhang, Joseph~E. Gonzalez, and Ion Stoica.
\newblock Judging llm-as-a-judge with mt-bench and chatbot arena.
\newblock In Alice Oh, Tristan Naumann, Amir Globerson, Kate Saenko, Moritz Hardt, and Sergey Levine (eds.), \emph{Advances in Neural Information Processing Systems 36: Annual Conference on Neural Information Processing Systems 2023, NeurIPS 2023, New Orleans, LA, USA, December 10 - 16, 2023}, 2023.
\newblock URL \url{http://papers.nips.cc/paper\_files/paper/2023/hash/91f18a1287b398d378ef22505bf41832-Abstract-Datasets\_and\_Benchmarks.html}.

\bibitem[Zheng et~al.(2024)Zheng, Wang, Wang, and An]{DBLP:conf/iclr/ZhengWW024}
Longtao Zheng, Rundong Wang, Xinrun Wang, and Bo~An.
\newblock Synapse: Trajectory-as-exemplar prompting with memory for computer control.
\newblock In \emph{The Twelfth International Conference on Learning Representations, {ICLR} 2024, Vienna, Austria, May 7-11, 2024}. OpenReview.net, 2024.
\newblock URL \url{https://openreview.net/forum?id=Pc8AU1aF5e}.

\bibitem[Zhou et~al.(2024)Zhou, Yan, Shlapentokh-Rothman, Wang, and Wang]{zhou2023lats}
Andy Zhou, Kai Yan, Michal Shlapentokh-Rothman, Haohan Wang, and Yu-Xiong Wang.
\newblock Language agent tree search unifies reasoning, acting, and planning in language models.
\newblock \emph{Proceedings of Machine Learning Research}, 235:\penalty0 62138--62160, 2024.

\bibitem[Zhou et~al.(2023{\natexlab{a}})Zhou, Sch{\"a}rli, Hou, Wei, Scales, Wang, et~al.]{zhou2023leasttomost}
Denny Zhou, Nathanael Sch{\"a}rli, Le~Hou, Jason Wei, Nathan Scales, Xuezhi Wang, et~al.
\newblock Least-to-most prompting enables complex reasoning in large language models.
\newblock In \emph{International Conference on Learning Representations}, 2023{\natexlab{a}}.

\bibitem[Zhou et~al.(2023{\natexlab{b}})Zhou, Xu, Zhu, Zhou, Lo, Sridhar, et~al.]{zhou2023webarena}
Shuyan Zhou, Frank~F. Xu, Hao Zhu, Xuhui Zhou, Robert Lo, Abishek Sridhar, et~al.
\newblock Webarena: A realistic web environment for building autonomous agents.
\newblock \emph{arXiv preprint arXiv:2307.13854}, 2023{\natexlab{b}}.

\end{thebibliography}


\end{document}